\documentclass[11pt,a4paper]{article}
\usepackage[hyperref]{naaclhlt2019}
\usepackage{times}
\usepackage{latexsym}
\pdfoutput=1

\usepackage{url}
\usepackage{booktabs}
\usepackage{multirow}

\usepackage{xspace}
\makeatletter
\DeclareRobustCommand\onedot{\futurelet\@let@token\@onedot}
\def\@onedot{\ifx\@let@token.\else.\null\fi\xspace}

\def\etal{et al\onedot}

\makeatother

\usepackage{amsmath}
\usepackage{amsfonts}

\aclfinalcopy 

\begin{document}
\newcommand\BibTeX{B{\sc ib}\TeX}

\title{Evaluation of Morphological Embeddings for English and Russian Languages}

\author{Vitaly Romanov \\
  Innopolis University, Innopolis, \\
  Russia \\
  {\tt v.romanov@innopolis.ru} \\\And
  Albina Khusainova \\
  Innopolis University, Innopolis, \\
  Russia \\
  {\tt a.khusainova@innopolis.ru} \\}

\date{}

\maketitle
\begin{abstract}
  This paper evaluates morphology-based embeddings for English and Russian languages. Despite the interest and introduction of several morphology-based word embedding models in the past and acclaimed performance improvements on word similarity and language modeling tasks, in our experiments, we did not observe any stable preference over two of our baseline models - SkipGram and FastText. The performance exhibited by morphological embeddings is the average of the two baselines mentioned above.
\end{abstract}

\section{Introduction}

One of the most significant shifts in the area of natural language processing is to the practical use of distributed word representations. \citet{collobert2011natural} showed that a neural model could achieve close to state-of-the-art results in Part of Speech (POS) tagging and chunking by relying almost only on word embeddings learned with a language model. In modern language processing architectures, high quality pre-trained representations of words are one of the major factors of the resulting model performance. 

Although word embeddings became ubiquitous, there is no single benchmark on evaluating their quality \cite{bakarov2018survey}, and popular intrinsic evaluation techniques are subject to criticism \cite{gladkova2016intrinsic}. Researchers very often rely on intrinsic evaluation, such as semantic similarity or analogy tasks. While intrinsic evaluations are simple to understand and conduct, they do not necessarily imply the quality of embeddings for all possible tasks \cite{bats2016}. 

In this paper, we turn to the evaluation of morphological embeddings for English and Russian languages. Over the last decade, many approaches tried to include subword information into word representations. Such approaches involve additional techniques that perform segmentation of a word into morphemes \cite{Arefyev, virpioja2013morfessor}. The presumption is that we can potentially increase the quality of distributional representations if we incorporate these segmentations into the language model (LM). 

Several approaches that include morphology into word embeddings were proposed, but the evaluation often does not compare proposed embedding methodologies with the most popular embedding vectors - Word2Vec, FastText, Glove. In this paper, we aim at answering the question of whether morphology-based embeddings can be useful, especially for languages with rich morphology (such as Russian). Our contribution is the following:

\begin{enumerate}
    \item We evaluate simple SkipGram-based (SG-based) morphological embedding models with new intrinsic evaluation BATS dataset \cite{bats2016}
    \item We compare relative gain of using morphological embeddings against Word2Vec and FastText for English and Russian languages
    \item We test morphological embeddings on several downstream tasks other than language modeling, i.e., mapping embedding spaces, POS tagging, and chunking
\end{enumerate}

The rest of the paper is organized as follows. Section \ref{sec:related_work} contains an overview of existing approaches for morphological embeddings and methods of their evaluation. Section \ref{sec:approaches} explains embedding models that we have tested. Section \ref{sec:evaluation} explains our evaluation approaches. Section \ref{sec:results} describes results.

\section{Related work} \label{sec:related_work}

The idea to include subword information into word representation is not new. The question is how does one obtain morphological segmentation of words. Very often, researchers rely on the unsupervised morphology mining tool Morfessor \cite{virpioja2013morfessor}.

Many approaches use simple composition, e.g., sum, of morpheme vectors to define a word embedding.  \citet{Botha2014} were one of the first to try this approach. They showed a considerable drop in perplexity of log-bilinear language model and also tested their model on word similarity and downstream translation task. The translation task was tested against an n-gram language model. Similarly, \citet{Qiu2014} tweak CBOW model so that besides central word it can predict target morphemes in this word. Final embeddings of morphemes are summed together into the word embedding. They test vectors on analogical reasoning and word similarity, showing that incorporating morphemes improves semantic similarity. \citet{El-kishky2018} develop their own morpheme segmentation algorithm and test the resulting embeddings on the LM task with SGNS objective. Their method achieved lower perplexity than FastText and SG. 

A slightly different approach was taken by \citet{Cotterell2015} who optimized a log-bilinear LM model with a multitask objective, where the second objective is to guess the next morphological tag. They test resulting vector similarity against string distance (morphologically close words have similar substrings) and find that their vectors surpass Word2Vec by a large margin.

\citet{Bhatia2016} construct a hierarchical graphical model that incorporates word morphology to predict the next word and then optimize the variational bound. They compare their model with Word2Vec and the one described by \citet{Botha2014}. They found that their method improves results on word similarity but is inferior to approach by \citet{Botha2014} in POS tagging.

Another group of methods tries to incorporate arbitrary morphological information into embedding model. \citet{Avraham2017} observe that it is impossible to achieve both high semantic and syntactic similarity on the Hebrew language. Instead of morphemes, they use other linguistic tags for the word, i.e., lemma, the word itself, and morphological tag. \citet{Chaudhary2018} took the next level of a similar approach. Besides including morphological tags, they include morphemes and character n-grams, and study the possibility of embedding transfer from Turkish to Uighur and from Hindi to Bengali. They test the result on NER and monolingual machine translation.

Another approach that deserves being mentioned here is FastText by \citet{bojanowski2017enriching}. They do not use morphemes explicitly, but instead rely on subword character n-grams, that store morphological information implicitly. This method achieves high scores on both semantic and syntactic similarities, and by far is the most popular word embedding model that also captures word morphology.

There are also approaches that investigate the impact of more complex models like RNN and LSTM. \citet{LuongThangandSocherRichardandManning2013} created a hierarchical language model that uses RNN to combine morphemes of a word to obtain a word representation. Their model performed well on word similarity task. Similarly, \citet{Cao2016} create Char2Vec BiLSTM for embedding words and train a language model with SG objective. Their model excels at the syntactic similarity.

\section{Embedding techniques} \label{sec:approaches}

In this work, we test three embedding models on English and Russian languages: SkipGram, FastText, and MorphGram. The latter one is similar to FastText with the only difference that instead of character n-grams we model word morphemes. This approach was often used in previous research.

All three models are trained using the negative sampling objective

\begin{multline}
\frac{1}{T} \sum_{t=1}^T \sum_{-m \leq j \leq m, j \neq 0} \log\sigma(s(w_j, w_t)) + \\
\sum_{i=1}^{k} \mathbb{E}_{w \sim P_n(w_t)}  \left[ \log\sigma(s(w, w_t)) \right] 
\end{multline}

In the case of SG, the similarity function \(s\) is the inner product of corresponding vectors. FastText and MorphGram are using subword units. We use the same approach to incorporate subword information into the word vector for both models:

\[
s(w_j, w_t) = \sum_{s\in \mathcal{S}_{w_t}} v_s^T v_{w_j}
\]
where \(S_{w_t}\) is the set of word segmentations into n-grams or morphemes. 
We use Gensim\footnote{\url{https://radimrehurek.com/gensim}} as the implementation for all models \cite{rehurek_lrec}. For MorphGram, we utilize FastText model and substitute the function that computes character n-grams for the function that performs morphological segmentation.

\section{Experiments and Evaluation} \label{sec:evaluation}

To understand the effect of using morphemes for training word embeddings, we performed intrinsic and extrinsic evaluations of SG, FastText, and MorphGram model for two languages - English and Russian. Russian language, in contrast to English, is characterized by rich morphology, which makes this pair of languages a good choice for exploring the difference in the effect of morphology-based models. 

\subsection{Data and Training Details}
We used the first 5GB of unpacked English and Russian Wikipedia dumps\footnote{\url{https://dumps.wikimedia.org/}} as training data. 

For training both SG and FastText we used Gensim library, for MorphGram - we adapted Gensim's implementation of FastText by breaking words into morphemes instead of n-grams, all other implementation details left unchanged. Training parameters remain the same as in the original FastText paper, except the learning rate was set to 0.05 at the beginning of the training, and vocabulary size was constrained to 100000 words. Morphemes for English words were generated with polyglot\footnote{\url{https://polyglot.readthedocs.io/en/latest/index.html}}, and for Russian - with seq2seq segmentation tool\footnote{\url{https://github.com/kpopov94/morpheme_seq2seq}}.

When reporting our results in tables, we will refer for FastText as FT and MorphGram as Morph.

\subsection{Similarity}
One of the intrinsic evaluations often used for word embeddings is a similarity test - given word pairs with human judgments of similarity degree for words in each pair, human judgments are compared with model scores---the more is the correlation, the better model ``understands" semantic similarity of words.  
We used SimLex-999 \cite{SimLex-999} dataset---the original one for English and its translated by \citet{leviant2015judgment} version for Russian, for evaluating trained embeddings. Out-of-vocabulary words were excluded from tests for all models. The results are presented in Table~\ref{simtable}.

\begin{table}
\footnotesize
\centering
\begin{center}
\begin{tabular}{llll}
\toprule
& \bfseries SG & \bfseries FT & \bfseries Morph  \\
\midrule
en & \textbf{0.37} & 0.35 & 0.36 \\ 
ru & \textbf{0.24} & 0.19 & 0.19 \\
\bottomrule
\end{tabular}
\end{center}
\caption{Correlation between human judgments and model scores for similarity datasets, Spearman's~$\rho$.}
\label{simtable}
\end{table}

We see that SG beats the other two models on similarity task for both languages, and MorphGram performs almost the same as Fasttext.

\subsection{Analogies}
Another type of intrinsic evaluations is analogies test, where the model is expected to answer questions of the form A is to B as C is to D, D should be predicted. For English, we used Google analogies dataset introduced by Mikolov \etal~\cite{mikolov2013efficient} and BATS collection \cite{bats2016}. For Russian, we used a partial translation\footnote{\url{https://rusvectores.org/static/testsets/}} of Mikolov's dataset, and a synthetic dataset by \citet{Abdou2018MGADMG}. 

Again, we excluded all out-of-vocabulary words from tests. We report accuracy for different models in Table~\ref{antable}.

\begin{table}
\footnotesize
\centering
\begin{center}
\begin{tabular}{lp{2.8cm}p{0.7cm}p{0.7cm}p{1cm}}
\toprule
&& \bfseries SG & \bfseries FT & \bfseries Morph  \\
\midrule
\multirow{3}{*}{en} & Google Semantic & \textbf{65.34} & 48.75 & 57.52 \\
& Google Syntactic & 55.88 & \textbf{75.10} & 61.16 \\ \cmidrule{2-5}
& BATS & 29.67 & \textbf{33.33} & 32.71 \\ \midrule
\multirow{3}{*}{ru} & Translated Semantic & \textbf{39.11} & 25.59 & 34.69 \\
& Translated Syntactic & 32.71 & \textbf{59.29} & 43.68 \\ \cmidrule{2-5}
& Synthetic & 24.52 & \textbf{36.78} & 27.06 \\
\bottomrule
\end{tabular}
\end{center}
\caption{Accuracy of models on different analogies tasks.}
\label{antable}
\end{table}

Interestingly, MorphGram is between SG and FastText in semantic categories for both languages, and between FastText and SG for syntactic categories for English.

\subsection{Mapping Embedding Spaces}
Here we introduce a new type of evaluation---it focuses on a cross-lingual task of mapping two embedding spaces for different languages. The core idea is to transform embedding spaces such that after this transformation the vectors of words in one language appear close to the vectors of their translations in another language. We were interested to see if using morphemes has any benefits to perform this kind of mapping. 

We map embeddings using a train seed dictionary (dictionary with word meanings) and state of the art supervised mapping method by \citet{Artetxe}, and calculate the accuracy of the mapping on the test dictionary. In short, the essence of this method is to find optimal orthogonal transforms for both embedding spaces to map them to a shared space based on a seed dictionary, plus some additional steps such as embeddings normalization. For each model---SG, FastText, and MorphGram, we mapped Russian and English embeddings trained using this model. We used the original implementation\footnote{\url{https://github.com/artetxem/vecmap}} for mapping (supervised option), and ground-truth train/test dictionaries provided by Facebook for their MUSE\footnote{\url{https://github.com/facebookresearch/MUSE}} library. We report 1-nn and 10-nn accuracy: whether the correct translation was found as a first nearest neighbor or among 10 nearest neighbors of a word in the mapped space. See the results in Table~\ref{maptable}.

\begin{table}
\footnotesize
\centering
\begin{center}
\begin{tabular}{llll}
\toprule
& \bfseries SG & \bfseries FT & \bfseries Morph  \\
\midrule
ru-en \textit{1-nn}  & \textbf{56.27} & 55.58 & 53.51 \\
ru-en \textit{10-nn}  & \textbf{78.96} & 78.82 & 77.03 \\ 
\bottomrule
\end{tabular}
\end{center}
\caption{Accuracy of supervised mapping from Russian to English using different models, searching among first and ten nearest neighbors.}
\label{maptable}
\end{table}

We observe no positive impact of using MorphGram for mapping word embedding spaces. 

\subsection{POS Tagging and Chunking}
Other tasks where incorporation of morphology can be crucial are the tasks of POS Tagging and chunking. We use a simple CNN-based architecture introduced in \cite{collobert2011natural}, with one projection layer, one convolutional layer, and the final logit layer. The only input features we use are the embeddings from corresponding models. The English language embeddings are tested with Conll2000 dataset which contains 8935 training sentences and 44 unique POS tags. The dataset for the Russian language contains 49136 sentences and 458 unique POS tags. Due to time constraint, we train models only for a fixed number of epochs: 50 for English and 20 for Russian (iterations reduced due to a larger training set). The results for POS and chunking are given in Tables \ref{tbl:pos} and \ref{tbl:chunk} correspondingly. It is interesting to note that SG embeddings perform better for English on POS task, but for Russian, embeddings that encode more syntactic information always perform better. 

\begin{table} 
\footnotesize
\centering
\begin{center}
\begin{tabular}{llll} 
\toprule
& \bfseries SG & \bfseries FT & \bfseries Morph  \\
\midrule
en & \textbf{0.9824} & 0.9754 & 0.9722 \\ 
ru & 0.8817 & \textbf{0.8899} & 0.8871 \\
\bottomrule
\end{tabular}
\end{center}
\caption{Accuracy on POS task}
\label{tbl:pos}
\end{table}

\begin{table} 
\footnotesize
\centering
\begin{center}
\begin{tabular}{llll} 
\toprule
& \bfseries SG & \bfseries FT & \bfseries Morph  \\
\midrule
en & 0.8966 & \textbf{0.9034} & 0.8985 \\ 
ru & 0.8442 & \textbf{0.8548} & 0.8534 \\
\bottomrule
\end{tabular}
\end{center}
\caption{Accuracy on Chunk task}
\label{tbl:chunk}
\end{table}

\section{Results} \label{sec:results}

In this paper, we compared three word embedding approaches for English and Russian languages. The main inquiry was about the relevance of providing morphological information to word embeddings. Experiments showed that morphology-based embeddings exhibit qualities intermediate between semantic driven embedding approaches as SkipGram and character-driven one as FastText. Morphological embeddings studied here showed average performance on both semantic and syntactic tests. We also studied the application of morphological embeddings on two downstream tasks: POS tagging and chunking. For English language, SG provided the best results for POS, whereas FastText gave the best result on chunking task. For Russian, FastText showed better performance on both tasks. Morphological embeddings, again, showed average results. We recognize that the difference in the results on downstream task can be considered marginal. We also did not observe improvements from morphological embeddings on word similarity dataset compared to other models.

\bibliography{naaclhlt2019}
\bibliographystyle{acl_natbib}

\end{document}